\documentclass{article}
\pdfoutput=1

\usepackage[nonatbib,preprint]{neurips_2020}

\usepackage[utf8]{inputenc} 
\usepackage[T1]{fontenc}    
\usepackage[hidelinks]{hyperref}       
\usepackage{url}            
\usepackage{booktabs}       
\usepackage{amsfonts}       
\usepackage{nicefrac}       
\usepackage{microtype}      

\usepackage{amsmath}

\usepackage{acro}
\usepackage{booktabs}
\usepackage[capitalise]{cleveref}
\usepackage{pgfplots}
\usepackage[all]{nowidow}
\usepackage{soul}
\usepackage{tabularx,multirow}
\usepackage{xspace}
\usepackage{tikz}
\usetikzlibrary{arrows,arrows.meta,patterns,positioning,shapes}

\newcommand{\axisunit}[1]{[#1]}

\DeclareAcronym{DDQL}{short=DDQL,long=double deep Q-learning}
\DeclareAcronym{NN}{short=NN,long=neural network}
\DeclareAcronym{RL}{short=RL,long=reinforcement learning}

\pgfdeclarelayer{background}
\pgfdeclarelayer{foreground}
\pgfsetlayers{background,main,foreground}

\pgfplotsset{compat=1.13,
	/pgfplots/ybar legend/.style={
		/pgfplots/legend image code/.code={%
			\draw[##1,/tikz/.cd,yshift=-0.25em]
			(0cm,0cm) rectangle (3pt,0.6em);},
	},
	scale only axis,
	xlabel near ticks,
	ylabel near ticks,
	xlabel style={font=\footnotesize,align=center},
	ylabel style={font=\footnotesize,align=center},
	xlabel style={yshift=1.5mm},
	ylabel style={yshift=-1.5mm},
	x tick label style={font=\footnotesize},
	y tick label style={font=\footnotesize},
	legend style={
		font={\footnotesize}
	},
	legend cell align={left},
	max space between ticks=20pt
}

\makeatletter
\@ifpackageloaded{algpseudocode}{
	\algrenewcommand\algorithmicrequire{\textbf{Input:}}
	\algrenewcommand\algorithmicensure{\textbf{Output:}}
	\algrenewcommand\algorithmiccomment[1]{{\small \hfill\(\triangleright\) #1}}

	\algnewcommand{\algorithmicbreak}{\textbf{break}}
	\algnewcommand{\Break}{\State \algorithmicbreak}
}{}
\makeatother

\tikzset{
	container/.style={inner sep=0pt, outer sep=0pt},
}

\tikzset{
	nnnode/.style={thick,draw,circle,inner sep=0pt,outer sep=0pt,minimum width=1cm},
	nnsynapse/.style={thick,-{Latex[width=2.5mm,length=3.5mm]}}
}
\newcommand{\nndummy}{
	\node[nnnode,fill=green!70!black] (l11) at (-2,1.25) {};
	\node[nnnode,fill=green!70!black] (l12) at (-2,-1.25) {};

	\node[nnnode,fill=blue!70!black] (l21) at (0,2.25) {};
	\node[nnnode,fill=blue!70!black] (l22) at (0,0.75) {};
	\node[nnnode,fill=blue!70!black] (l23) at (0,-0.75) {};
	\node[nnnode,fill=blue!70!black] (l24) at (0,-2.25) {};

	\node[nnnode,fill=purple!70!white] (l31) at (2,0) {};

	\draw[nnsynapse] (l11) to (l21);
	\draw[nnsynapse] (l11) to (l22);
	\draw[nnsynapse] (l11) to (l23);
	\draw[nnsynapse] (l11) to (l24);
	\draw[nnsynapse] (l12) to (l21);
	\draw[nnsynapse] (l12) to (l22);
	\draw[nnsynapse] (l12) to (l23);
	\draw[nnsynapse] (l12) to (l24);
	\draw[nnsynapse] (l21) to (l31);
	\draw[nnsynapse] (l22) to (l31);
	\draw[nnsynapse] (l23) to (l31);
	\draw[nnsynapse] (l24) to (l31);
}
\newcommand{\largenndummy}{
	\node[nnnode,fill=green!70!black] (l11) at (-2,1.25) {};
	\node[nnnode,fill=green!70!black] (l12) at (-2,-1.25) {};

	\node[nnnode,fill=blue!70!black] (l21) at (0,2.25) {};
	\node[nnnode,fill=blue!70!black] (l22) at (0,0.75) {};
	\node[nnnode,fill=blue!70!black] (l23) at (0,-0.75) {};
	\node[nnnode,fill=blue!70!black] (l24) at (0,-2.25) {};

	\node[nnnode,fill=orange!70!white] (l31) at (2,1.5) {};
	\node[nnnode,fill=orange!70!white] (l32) at (2,0) {};
	\node[nnnode,fill=orange!70!white] (l33) at (2,-1.5) {};

	\node[nnnode,fill=purple!70!white] (l41) at (4,0) {};

	\draw[nnsynapse] (l11) to (l21);
	\draw[nnsynapse] (l11) to (l22);
	\draw[nnsynapse] (l11) to (l23);
	\draw[nnsynapse] (l11) to (l24);
	\draw[nnsynapse] (l12) to (l21);
	\draw[nnsynapse] (l12) to (l22);
	\draw[nnsynapse] (l12) to (l23);
	\draw[nnsynapse] (l12) to (l24);

	\draw[nnsynapse] (l21) to (l31);
	\draw[nnsynapse] (l22) to (l31);
	\draw[nnsynapse] (l23) to (l31);
	\draw[nnsynapse] (l21) to (l32);
	\draw[nnsynapse] (l22) to (l32);
	\draw[nnsynapse] (l23) to (l32);
	\draw[nnsynapse] (l24) to (l32);
	\draw[nnsynapse] (l22) to (l33);
	\draw[nnsynapse] (l23) to (l33);
	\draw[nnsynapse] (l24) to (l33);

	\draw[nnsynapse] (l31) to (l41);
	\draw[nnsynapse] (l32) to (l41);
	\draw[nnsynapse] (l33) to (l41);
}
\newcommand{\nnnode}[3]{
	\node[container,#2] (#1) {
		\resizebox{!}{#3}{
			\begin{tikzpicture}
			\nndummy
			\end{tikzpicture}
		}
	};
}
\newcommand{\largennnode}[3]{
	\node[container,#2] (#1) {
		\resizebox{!}{#3}{
			\begin{tikzpicture}
			\largenndummy
			\end{tikzpicture}
		}
	};
}

\newcolumntype{L}[1]{>{\raggedright\let\newline\\\arraybackslash\hspace{0pt}}m{#1}}
\newcolumntype{C}[1]{>{\centering\let\newline\\\arraybackslash\hspace{0pt}}m{#1}}
\newcolumntype{R}[1]{>{\raggedleft\let\newline\\\arraybackslash\hspace{0pt}}m{#1}}

\makeatother

\AtBeginDocument{
}

\usepgfplotslibrary{fillbetween}
\usetikzlibrary{calc}

\usepackage{algorithm}
\usepackage[noend]{algpseudocode}
\usepackage{xstring}
\usepackage{xparse}

\usepackage[nearskip=2pt]{subfig}

\title{Distributed Learning on\\Heterogeneous Resource-Constrained Devices}

\author{%
  Martin Rapp$^1$, Ramin Khalili$^2$, J\"org Henkel$^1$ \\
  $^1$Karlsruhe Institute of Technology, Karslruhe, Germany\\
  $^2$Huawei Research Center, Munich, Germany \\
  \texttt{martin.rapp@kit.edu},\ \ \ \texttt{ramin.khalili@huawei.com},\ \ \  \texttt{henkel@kit.edu}
}

\begin{document}

\maketitle

\begin{abstract}
We consider a distributed system, consisting of a heterogeneous set of devices, ranging from low-end to high-end.
These devices have different profiles, e.g., different energy budgets, or different hardware specifications, determining their capabilities on performing certain learning tasks. 
We propose the first approach that enables distributed learning in such a heterogeneous system.
Applying our approach, each device employs a \ac{NN} with a topology that fits its capabilities; however, part of these \acp{NN} share the same topology, so that their parameters can be jointly learned.
This differs from current approaches, such as federated learning, which require all devices to employ the same \ac{NN}, enforcing a trade-off between achievable accuracy and computational overhead of training.
We evaluate heterogeneous distributed learning for \ac{RL} and observe that it greatly improves the achievable reward on more powerful devices, compared to current approaches, while still maintaining a high reward on the weaker devices.
We also explore supervised learning, observing similar gains.
\end{abstract}

\acresetall

\def\xmax{100}

\pgfplotsset{
	isolated/.style={
		purple, densely dotted,
	},
	homogeneous/.style={
		blue, densely dashed,
	},
	heterogeneous/.style={
		green!50!black,
	},
	legend image code/.code={
		\draw[mark repeat=2,mark phase=2,#1,draw=none,fill,opacity=0.3]
			(0cm,-0.1cm) rectangle 
			(0.6cm,0.1cm);
		\draw[mark repeat=2,mark phase=2,#1]
			plot coordinates {
				(0cm,0cm)
				(0.3cm,0cm)
				(0.6cm,0cm)
			};
	},
}

\newcommand{\drawstd}[3]{
	\addplot[#1,very thick] table[x expr=\xmax*\thisrow{frac},y expr=100*\thisrow{#3_median},col sep=comma] {#2};
	
	\addplot[forget plot,draw=none,name path=lo] table[x expr=\xmax*\thisrow{frac},y expr=100*\thisrow{#3_min},col sep=comma] {#2};
	\addplot[forget plot,draw=none,name path=hi] table[x expr=\xmax*\thisrow{frac},y expr=100*\thisrow{#3_max},col sep=comma] {#2};
	\addplot[forget plot,#1,draw=none,opacity=0.3] fill between[of=lo and hi];
}

\newcommand{\supervisedlearningcurve}[3]{
	\addplot[#1,very thick] table[x expr=\thisrow{iteration}/1000,y expr=100*\thisrow{#3_median},col sep=comma] {#2};
	
	\addplot[forget plot,draw=none,name path=lo] table[x expr=\thisrow{iteration}/1000,y expr=100*\thisrow{#3_min},col sep=comma] {#2};
	\addplot[forget plot,draw=none,name path=hi] table[x expr=\thisrow{iteration}/1000,y expr=100*\thisrow{#3_max},col sep=comma] {#2};
	\addplot[forget plot,#1,draw=none,opacity=0.3] fill between[of=lo and hi];
}

\newcommand{\ataricurve}[3]{
	\IfFileExists{#2}{
		\addplot[#1,very thick] table[x expr=\thisrow{iteration}/100,y=#3_median,col sep=comma] {#2};
	
		\addplot[forget plot,draw=none,name path=lo] table[x expr=\thisrow{iteration}/100,y expr=\thisrow{#3_min},col sep=comma] {#2};
		\addplot[forget plot,draw=none,name path=hi] table[x expr=\thisrow{iteration}/100,y expr=\thisrow{#3_max},col sep=comma] {#2};
		\addplot[forget plot,#1,draw=none,opacity=0.3] fill between[of=lo and hi];
	}{}
}
\section{Introduction}
\label{sec:intro}

Recently, techniques for distributed learning, such as \emph{federated learning}~\cite{federated} and \emph{A3C}~\cite{a3c}, have been proposed.
Such techniques allow the training of \acp{NN} to be performed in a distributed manner, where every device trains a local copy of the \ac{NN} with locally-collected/available training data, and parameter deltas are exchanged among devices, by the help of a coordinator, for synchronization.
Distributed learning has proven effective in large-scale systems~\cite{google_keyboard}.
However, to apply these techniques, all devices have to use and train the same~\ac{NN}, limiting their applicability.

In particular, there is a trade-off between the quality of a model and its resource requirements: to achieve a high level of accuracy, we need to employ complex models, which require a large level of resources~\cite{nn_size_tradeoff}.
However, distributed systems often consist of a heterogeneous set of devices with differing constraints in terms of computational capability (e.g., through the presence of an accelerator), available memory, power (e.g., because of thermal constraints), energy (e.g., for battery-driven devices), etc.
These constraints put upper bounds on the achievable accuracy, e.g., energy consumption strongly correlates with the achievable accuracy~\cite{nn_size_tradeoff}.
An example of such systems is a distributed wireless sensor network, composed of a number of sensor nodes, each employing an \ac{NN} model to detect events.
Power supplies based on energy harvesting, such as solar cells, are subject to variations, which leads to heterogeneous energy and power budgets for devices~\cite{shaikh2016energy}.
Another example is a system of smartphones, each using an \ac{NN} model to predict the user's next action.
Smartphones have different computational capabilities, as some may have a GPU or even an \ac{NN} accelerator~\cite{aibenchmark}.

Existing distributed learning techniques cannot cope with such heterogeneity.
They are forced to select the \ac{NN} topology based on the constraints of the weakest devices, i.e., employing a lightweight model, reducing the accuracy level that otherwise could be achieved in more powerful devices by deploying a more complex \ac{NN}.
One may propose to split the devices into several groups, according to their capabilities, employing different \acp{NN} for each group.
As these groups do not share a common \ac{NN}, synchronization of their \ac{NN} parameters is not possible.
Hence, devices from one group cannot benefit from training data collected in another group, failing to gain from cooperative learning. 

In this work, we propose a novel approach to enable cooperative learning in distributed and heterogeneous systems.
Applying our solution, each device uses a topology that fits its constraints, hence, we would have \acp{NN} of different complexities employed by the devices, i.e., a complex (resp.\ lightweight) \ac{NN} on powerful (resp.\ weak) devices.
However, parts of these \acp{NN} share the same topology -- enabling synchronizing their parameters.
Thereby, cooperative learning between these devices is achieved.
We demonstrate our technique for distributed \ac{RL}, when all \acp{NN} share the same topology only for the first few layers, and when the parameters of these shared layers are merged using a simple averaging.
We observe that cooperative learning greatly improves the achievable reward on more powerful devices, compared to current approaches, while still maintaining a high reward on the weaker devices. 

We also explore distributed supervised learning, showing that by applying some adjustments on how the layers are shared among the devices, and how the parameters of different devices are merged, similar gains can be achieved.
Note that in a distributed system, the training data is another resource that is heterogeneously distributed among devices.
This is as i) different devices could have different capacity on collecting and storing data, and ii), in a distributed system, data is collected locally at each device, whereas due to communication and privacy constraints, the collected data cannot be fully shared with other devices.
We show that in the case of supervised learning, this heterogeneity needs to be also taken into account when merging the parameters of the shared layers.

\section{Related work}
\label{sec:related}

Many studies focus on \emph{parallel} learning~\cite{povey2014parallel,lian2015asynchronous}, where training data is gathered on a central coordinator and is then distributed to all computing devices.
In contrast, \emph{distributed} learning considers the case when data is collected and processed locally on each device.
Distributed learning can be divided into synchronous and asynchronous methods.
Federated learning~\cite{federated} uses synchronous update rounds, in which first, the current \ac{NN} parameters are distributed to all devices by a coordinator.
Then, each device performs a batch of training to obtain an updated set of parameters.
Finally, each device synchronously sends its updated parameters back to the coordinator, where the average of all parameters is calculated and the cycle repeats.
Variations of this technique are proposed to take into account privacy~\cite{privacy_deep_learning} or communication constraints~\cite{federated_communication}.
Furthermore, asynchronous updates have also been shown effective~\cite{a3c}.
However, these works do not consider heterogeneity in the devices and assume that all devices are equally capable of training an~\ac{NN} and, therefore, unlike our work, employ the same \ac{NN} on every device.
The work in~\cite{li2018federated} extends federated learning towards heterogeneous systems by allowing devices to calculate incomplete parameter updates if they run out of time.
However, still, all devices have to be capable of training the same \ac{NN}.
Furthermore, only a subset of the training data may be used on weak devices.

The trade-off between an accurate but complex \ac{NN}, and a lightweight but inaccurate \ac{NN} has been exploited in many works to cope with resource constraints.
Authors of~\cite{park2015big} employ two \acp{NN} to speed up the inference.
Inference always starts with the lightweight \ac{NN}.
Only if it reports a low confidence, the complex \ac{NN} is used.
The topologies of the two \acp{NN} are independent, i.e., there is no shared part, and the \acp{NN} are trained independently.
The work in~\cite{samie2020hierarchical} uses an additional lightweight classifier to decide which \ac{NN} to use.
The work in~\cite{li2015convolutional} proposes to build a cascade of \acp{NN}, in which the lightweight \ac{NN} is a part of the complex \ac{NN}.
When building the \ac{NN}, first the lightweight \ac{NN} is trained.
Its parameters are fixed during the training of the complex \ac{NN}.
The work in~\cite{leroux2015resource} builds a cascade of many classifiers that share initial layers.
They first train a complex \ac{NN} and fix its parameters.
Then, they add a single output layer after \emph{every} hidden layer and train it.
All these techniques make the decision whether to use the lightweight or complex \ac{NN} depending on the sample.
This is in contrast to our work, where devices have different resource constraints that determine which \ac{NN} can be employed, and where these different \acp{NN} are trained in parallel and cooperatively.

Some studies consider a different type of heterogeneous learning where not the training data but the features are distributed~\cite{ying2018supervised}, i.e., each device observes a part of the feature vector.
Furthermore, our proposed technique bears similarities to transfer learning~\cite{transfer-learning}.
However, transfer learning considers the use case when a single \ac{NN} is retrained on a different task than it was originally trained for by reusing learned parameters.
We consider the use case when different \acp{NN} are jointly trained on the same task.
Finally, a plethora of works considers \ac{NN} training and inference under resource constraints and propose many techniques such as weight quantization~\cite{hubara2017quantized} or pruning~\cite{molchanov2016pruning}.
These works do not consider heterogeneity across devices and thus differ from our work.

\section{Distributed learning under heterogeneous resource constraints}
\label{sec:distributed_learning}

\newcommand{\devicegroup}[2]{
	\foreach \i in {1,2,3} {
		\node[draw,inner sep=0pt,outer sep=0pt,minimum width=#1,minimum height=1cm,anchor=center] (dev\i) at (1.5*\i,0) {};
		#2{}{above=1mm of dev\i.south,anchor=south}{0.5cm}
		\node[font=\footnotesize\strut,inner sep=1pt,anchor=center] (env\i) at (1.5*\i,-1.2) {Env.};
		\node[font=\footnotesize\strut,inner sep=1pt,anchor=north] at (dev\i.north) {Dev.};

		\draw[-latex',bend left] ([xshift=+1mm]dev\i.south) to ([xshift=+1mm]env\i.north);
		\draw[-latex',bend left] ([xshift=-1mm]env\i.north) to ([xshift=-1mm]dev\i.south);
	}

	\draw[latex'-latex',bend left=15] ([xshift=1mm,yshift=6mm]dev1.center) to ([xshift=-1mm,yshift=6mm]dev2.center);
	\draw[latex'-latex',bend left=15] ([xshift=1mm,yshift=6mm]dev2.center) to ([xshift=-1mm,yshift=6mm]dev3.center);
	\draw[latex'-latex',bend left=15] ([xshift=1mm,yshift=6mm]dev1.center) to node [above,font=\footnotesize\strut,inner sep=1pt,align=center] {state-of-the-art\\[-2pt]homogeneous synchronization} ([xshift=-1mm,yshift=6mm]dev3.center);
}

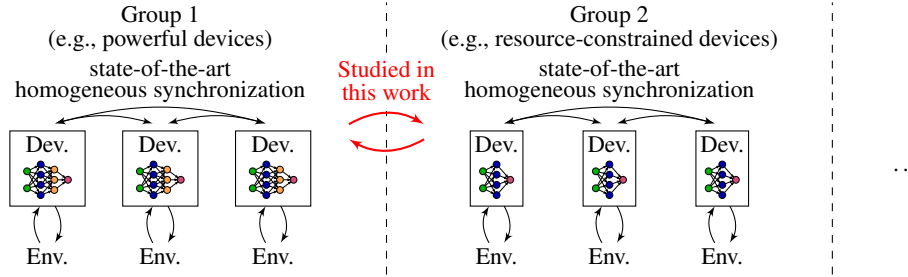
\begin{figure}
	\centering
	\begin{tikzpicture}
		\node[container,anchor=south east] (large) at (-1.0,0) {
			\begin{tikzpicture}
				\devicegroup{1cm}{\largennnode}
			\end{tikzpicture}
		};
		\node[font=\footnotesize\strut,inner sep=1pt,align=center,anchor=south] (largelabel) at (large.north |- {(0,2.9)}) {Group 1\\[-1pt](e.g., powerful devices)};

		\node[container,anchor=south west] (small) at (+1.0,0) {
			\begin{tikzpicture}
				\devicegroup{0.7cm}{\nnnode}
			\end{tikzpicture}
		};
		\node[font=\footnotesize\strut,inner sep=1pt,align=center,anchor=south] (smalllabel) at (small.north |- {(0,2.9)}) {Group 2\\[-1pt](e.g., resource-constrained devices)};

		\draw[dashed] ({(0,0)} |- large.south) to ({(0,0)} |- largelabel.north);

		\draw[red,thick,-latex',bend left] (-0.5,2.0) to node [above=0.5pt,font=\footnotesize\strut,align=center,inner sep=1pt,fill=white] {Studied in\\[-1pt]this work} ++(+1cm,0);
		\draw[red,thick,-latex',bend left] (+0.5,1.8) to ++(-1cm,0);

		\draw[dashed] ([xshift=1cm]small.east |- large.south) to ([xshift=1cm]small.east |- largelabel.north);
		\node[font=\footnotesize\strut] at ([xshift=2cm]small.east) {\ldots};
	\end{tikzpicture}
	\caption{The application model underlying this work.}
	\label{fig:model}
\end{figure}

\cref{fig:model} summarizes our application model: a heterogeneous set of devices, where each device interacts with its environment using an \ac{NN}.
During the interaction, training data is collected on every device, which will be used by the device to train and improve the \ac{NN}.
The training is performed locally on each device, using the locally-collected data, where reinforcement learning or supervised learning techniques can be used.
We assume that the devices are interacting with \textit{similar} environments, e.g., in the case of reinforcement learning, they share the same underlying (Hidden) Markov model but are independent random instantiations of this model.
This enables cooperative learning among all devices, in which each device can benefit from experiences gathered and learned by all other devices.
However, devices are subject to heterogeneous resource constraints.

In particular, there are two types of resource constraints: those that determine the computational and communicational capabilities of devices, and the amount of training data available on each device.
A straightforward constraint for computational capabilities is the peak performance (e.g., FLOPS) that is achievable on the device.
It is determined by factors such as processor frequency and microarchitecture, or the availability of accelerators like a GPU or a neural accelerator~\cite{aibenchmark}.
The amount of memory or storage on a device limits the number of parameters that an \ac{NN} can deploy on the device.
Limited communication bandwidth bounds the rate and amount of data exchange between devices.
Power, energy, and temperature constraints limit the instantaneous or average performance on the device, as well as the amount of data that can be transmitted to or received from others.

These constraints affect distributed learning as follows.
Limited computation capabilities (peak performance, power, energy, temperature) put a limit on the number of operations that a device can perform for training and inference.
The amount of parameters of the \ac{NN} is bounded by both the available memory and storage on the device, applying limitations on the complexity of the employed \ac{NN}.
Communication constraints limit the number and rate of parameter updates that are exchanged between devices.
Finally, the available training data may vary among devices, as devices might have a different rate of interaction with the environment, and thus collecting training data at different rates, or have different storage capacity to store the collected data.
\emph{Most importantly, constraints vary between different devices -- forming a heterogeneous system.}

\emph{Objective:}
The objective followed in this work is to maximize the achievable reward / accuracy on all devices, given that devices have different resource profiles.

\emph{Model:}
To exploit the available resources on each device, we group all devices with similar resource profiles.
Each of these groups employs an \ac{NN} model that fits its constraints.
Synchronizing \ac{NN} parameters within one group can be done with state-of-the-art techniques such as conventional federated learning since all devices employ the same \ac{NN} model and have a similar amount of training data.
This allows us to treat each group of devices as if it were a single device.
We don't study how to select these \acp{NN} on different devices, but rather use a simple abstracted model, where we assume different categories of devices, each with a certain resource profile, capable of employing a certain \ac{NN} with a certain complexity.

\emph{Challenges:}
The challenge tackled in this work is how to synchronize \ac{NN} parameters across devices of different groups, as for each group the employed \ac{NN} model and the amount of training data varies.
The goal is to enable cooperative learning across all devices.

\emph{Limitations:} 
We chose to abstract from factors such as communication latency or unreliable communication channels to be able to focus on convergence properties.

\paragraph{Our approach (heterogeneous learning):}

\tikzset{
	label/.style={font=\footnotesize\vphantom{Ag},align=center,anchor=center,inner sep=4pt},
	nnlayer/.style={inner sep=0pt,draw,black,fill=white,anchor=center,minimum width=0.25cm,minimum height=2cm},
}

\newcommand{\nnconnect}[2]{
	\draw[black,-latex'] ([yshift=-1mm]#1.north east) to ([yshift=-1mm]#2.north west);
	\draw[black,-latex'] ([yshift=1mm]#1.south east) to ([yshift=1mm]#2.south west);
	\draw[black,-latex'] ([yshift=-2mm]#1.north east) to ([yshift=2mm]#2.south west);
	\draw[black,-latex'] ([yshift=2mm]#1.south east) to ([yshift=-2mm]#2.north west);
}

\begin{figure}
	\centering
	\begin{tikzpicture}
		\node[nnlayer,minimum height=1.5cm] (l1) at (0,0) {};
		\node[nnlayer,minimum height=1.75cm] (l2) at (1,0) {};
		\node[nnlayer,minimum height=1.25cm] (l31) at (2,0.8) {};
		\node[nnlayer,minimum height=1.1cm] (l41) at (3,0.8) {};
		\node[nnlayer,minimum height=0.75cm] (l51) at (4,0.8) {};

		\node[nnlayer,minimum height=1cm] (l32) at (2,-0.8) {};
		\node[nnlayer,minimum height=0.75cm] (l42) at (3,-0.8) {};

		\nnconnect{l1}{l2}
		\nnconnect{l2}{l31}
		\nnconnect{l2}{l32}
		\nnconnect{l31}{l41}
		\nnconnect{l41}{l51}
		\nnconnect{l32}{l42}

		\draw [blue,thick,densely dashed] plot [smooth cycle,tension=0.5] coordinates {
			([xshift=-0.5mm,yshift=0.5mm]l1.north west)
			([xshift=-0.5mm,yshift=0.5mm]l2.north west)
			([xshift=-0.5mm,yshift=0.5mm]l31.north west)
			([yshift=0.5mm]l41.north east)
			([xshift=0.5mm,yshift=0.5mm]l51.north east)
			([xshift=2mm]l51.east)
			([xshift=0.5mm,yshift=-0.5mm]l51.south east)
			([yshift=-0.5mm]l41.south east)
			([xshift=0.5mm,yshift=-0.5mm]l31.south east)
			([xshift=-0.5mm,yshift=0.5mm]l32.north west)
			([xshift=0.5mm,yshift=-0.5mm]l2.south east)
			([xshift=-0.5mm,yshift=-0.5mm]l1.south west)
		};
		\draw [red,thick,densely dotted] plot [smooth cycle,tension=0.5] coordinates {
			([xshift=-0.5mm,yshift=0.5mm]l1.north west)
			([xshift=0.5mm,yshift=0.5mm]l2.north east)
			([xshift=-0.5mm,yshift=-0.5mm]l31.south west)
			([xshift=0.5mm,yshift=0.5mm]l32.north east)
			([xshift=0.5mm,yshift=0.5mm]l42.north east)
			([xshift=2mm]l42.east)
			([xshift=0.5mm,yshift=-0.5mm]l42.south east)
			([yshift=-0.5mm]l32.south west)
			([xshift=-0.5mm,yshift=-0.5mm]l2.south west)
			([xshift=-0.5mm,yshift=-0.5mm]l1.south west)
		};

		\draw[latex'-] (l1.west) to ++(-5mm,0) node[label,left] {Input};
		\draw[-latex'] (l51.east) to ++(5mm,0) node[label,right] {High accuracy output};
		\draw[-latex'] (l42.east) to ++(5mm,0) node[label,right] {Low accuracy output};

		\node [label,red,align=left,anchor=west] at ([xshift=1.5mm,yshift=-5mm]l42.east) {Lightweight branch: low resource requirements};
		\node [label,blue,align=left,anchor=west] at ([xshift=3mm,yshift=5mm]l51.east) {Complex branch: high resource requirements};
	\end{tikzpicture}
	\caption{
		In our proposed heterogeneous distributed learning, each device employs an \ac{NN} with different complexity (and accuracy) according to its resources.
		The first layers of all networks share the topology and parameters.
		The following layers are selected based on different trade-offs between resource requirements and accuracy.
		This figure shows the resulting topology for two types of devices.
	}
	\label{fig:merging}
\end{figure}
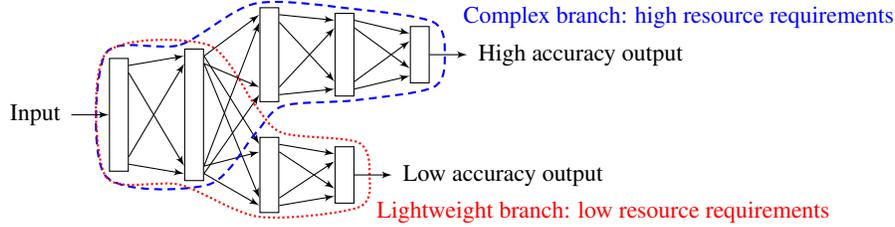

\cref{fig:merging}~illustrates and example on how distributed learning under in a heterogeneous distributed system is possible.
As explained before, we employ on each device the most accurate, and, therefore, most complex, \ac{NN} that fits its constraints.
However, to still enable cooperative learning, all \ac{NN} share the first few layers of their topology.
This enables synchronizing their parameters.
A positive side-effect of synchronizing only the first few layers is that the number of parameter updates is also reduced, mitigating communication constraints.
An alternative view on this architecture is that there exists one global topology with several branches.
Every device uses only a single branch for training and inference, and, therefore, only needs to store and update its parameters.
The stem of this NN (first few layers) is stored, updated, and synchronized by all devices, which enables cooperative learning.

\begin{figure}
	\centering
	\begin{minipage}{0.8\linewidth}
		\begin{algorithm}[H]
			\caption{Heterogeneous distributed learning on each device $k$}
			\footnotesize
			\label{algo:devices}
			\begin{algorithmic}
				\State clear gradient: $d\theta \gets 0$
				\State init.\ NN parameters: $\theta_{local} \gets$ random initialization;~~receive $\theta_{shared}$ from coordinator
				\State concat $\theta \gets (\theta_{local}, \theta_{shared})$
				\For{\textbf{each} time step $t$}
					\State interact with environment and collect data
					\State sample mini-batch $B$ for training from local memory
					\State accumulate gradient $d\theta \gets d\theta + \frac{\partial}{\partial\theta} L(B;\theta)$ w.r.t. loss function $L$
					
					\If {$t\ \mathrm{mod}\ T = 0$}
						\State split $(d\theta_{shared}, d\theta_{local}) \gets d\theta$
						\State send $d\theta_{shared}$ to coordinator as $d\theta_{shared}^{(k)}$
						\State receive new $\theta_{shared}$ from coordinator
						\State update local parameters $\theta_{local}$ based on the gradient $d\theta_{local}$
						\State concat $\theta \gets (\theta_{local}, \theta_{shared})$
						\State clear gradient: $d\theta \gets 0$
					\EndIf
				\EndFor
			\end{algorithmic}
		\end{algorithm}
	\end{minipage}
\end{figure}
\begin{figure}
	\begin{minipage}{0.45\linewidth}
		\begin{algorithm}[H]
			\caption{Synchronous coordinator}
			\footnotesize
			\label{algo:syncserver}
			\begin{algorithmic}
				\State $\theta_{shared} \gets$ random initialization
				\While{true}
					\State broadcast $\theta_{shared}$ to all devices
					\State receive all $d\theta_{shared}^{(k)}$
					\State $d\theta_{shared} \gets \sum_k d\theta_{shared}^{(k)}$
					\State $\theta_{shared} \gets \theta_{shared} + d\theta_{shared}$
				\EndWhile
			\end{algorithmic}
		\end{algorithm}
	\end{minipage}
	\hfill
	\footnotesize or
	\hfill
	\begin{minipage}{0.45\linewidth}
		\begin{algorithm}[H]
			\caption{Asynchronous coordinator}
			\footnotesize
			\label{algo:asyncserver}
			\begin{algorithmic}
				\State $\theta_{shared} \gets$ random initialization
				\State broadcast $\theta_{shared}$ to all devices
				\While{true}
					\State receive any $d\theta_{shared}^{(k)}$
					\State $\theta_{shared} \gets \theta_{shared} + d\theta_{shared}^{(k)}$
					\State send $\theta_{shared}$ to device~$k$
				\EndWhile
			\end{algorithmic}
		\end{algorithm}
	\end{minipage}
\end{figure}

\cref{algo:devices} shows pseudocode of how synchronization is performed on the devices.
In each time step, the devices interact with their environment to collect training data, and perform an update step of their local model.
Every $T$ time steps (one round), devices synchronize the accumulated gradients of the shared layers with the help of the coordinator.
Note that we abstract from many implementation details to keep the technique general.
The key difference to state-of-the-art techniques is that each device only sends a fraction of its parameter updates, $d\theta_{shared}$, to the coordinator, which leads to only this fraction being synchronized among devices.
This technique is orthogonal to the parameter exchange algorithm (e.g., asynchronous or synchronous updates) and, therefore, is fully compatible with existing distributed learning techniques.
\cref{algo:syncserver,algo:asyncserver} list example implementations of a synchronous and asynchronous coordinator based on synchronous \emph{FedAvg}~\cite{federated} and asynchronous coordination proposed in~\cite{a3c}, respectively.

\section{Experiments}
\label{sec:experiments}

This section presents empirical evaluations of our solution for distributed \ac{RL} and supervised learning tasks.
We first introduce the baselines. 

\newcommand{\powerful}{{powerful}\xspace}
\newcommand{\weak}{{weak}\xspace}
\newcommand{\Powerful}{{Powerful}\xspace}
\newcommand{\Weak}{{Weak}\xspace}

\subsection{Baselines}
\label{sec:baselines}

\paragraph{Isolated learning:}
A simple baseline is where each device employs the most accurate \ac{NN} topology that fits its available resources.
This results in a complex (resp.\ lightweight) \ac{NN} being employed on powerful (resp.\ weak) devices.
All devices train their local \ac{NN} with their locally-available training data.
There is no synchronization between the devices and, therefore, no cooperative learning.

\paragraph{Homogeneous distributed learning (state-of-the-art):}
The second baseline follows the state of the art on distributed learning.
All devices cooperatively train the same \ac{NN} with their locally-available training data.
For an \ac{RL} task, this follows the technique proposed in~\cite{a3c}.
For a supervised learning task, this closely models \emph{federated learning}~\cite{federated}.
As all devices train the same network, the \ac{NN} topology is limited by the weakest device, which results in a lightweight model being employed.
\subsection{Reinforcement learning}
\label{sec:rl}

We choose the \emph{Atari 2600} game suite~\cite{atari} as a benchmark for distributed \ac{RL}.
We study how different learning algorithms (isolated, homogeneous, heterogeneous) perform in a heterogeneous setup, reporting the first 10 million steps of training.
Our goal is not to study the absolute performance we can achieve using our solution, but rather its feasibility and its relative gain compared to the baselines.
Hence, for the sake of simplicity in the implementation, we select a setup based on \ac{DDQL} with experience replay~\cite{rl_atari}.
More advanced solutions, such as A3C algorithm~\cite{a3c}, could be studied in future work, but we expect to observe a similar behavior.

We consider a scenario composed of two devices with different resource constraints, denoted by \emph{\powerful} and \emph{\weak} devices.
Correspondingly, we employ two \ac{NN} topologies with different resource requirements, a \emph{complex} and a \emph{lightweight} \ac{NN}. They share the first few layers as depicted in \cref{fig:merging}.
The parameters of the complex \ac{NN} are based on~\cite{rl_atari}.
We also select the training parameters, replay window, etc.\ based on~\cite{rl_atari}.
The lightweight \ac{NN} has a reduced number of channels in the last convolutional and fully-connected layers, reducing the number of computational operations and parameters to 66\,\% and 4.2\,\%, respectively.
The number of computational operations is a good estimate of energy consumption, whereas the number of parameters is a good proxy for memory requirements~\cite{nn_size_tradeoff}.
Furthermore, we reduce the size of the experience replay buffer used in combination with the lightweight \ac{NN} to 10\,\% compared to the complex \ac{NN}.

Synchronization between devices is performed periodically.
\Ac{DDQL} already periodically copies the parameters from the updated \ac{NN} to the target \ac{NN}.
The synchronization between devices is performed at the same time, 
with the synchronized updated parameters being copied to the target network.
Homogeneous learning synchronizes all \ac{NN} parameters, heterogeneous learning only synchronizes the parameters of the first few shared layers, and isolated learning does not synchronize any parameters.
We perform training on each device at the same rate (one mini-batch per interaction with the environment).
We implement a synchronous coordinator that is invoked every 10,000~steps.
Before synchronization, we introduce a single test epoch on every device to measure its current performance.
Therefore, we temporarily reduce the $\epsilon$ of the $\epsilon$-greedy policy to $0.02$.
Also, as explained in~\cref{sec:distributed_learning}, the rate of interaction of the devices with their environments could be different.
Therefore, we perform additional experiments in which the \weak device interacts with its environment, and gathers training data, at a rate~$v_{weak}$ of $0.5{\times}$ and $0.25{\times}$ of the \powerful device.  

\begin{figure}
	\centering
	\pgfplotsset{
		rlaxis/.style={
			anchor=north west,
			width=2.7cm,
			height=2.05cm,
			xmin=0,
			xmax=10,
			xtick distance=2,
			ymin=0,
			xmajorgrids,
			ymajorgrids,
			xlabel={Million steps},
			ylabel={Test ep. reward},
			legend style={
				at={(1.0,1.0)},
				anchor=south,
				legend columns=3,
				xshift=0.6cm,
				yshift=0.7cm,
				fill=none,
			},
			legend style={/tikz/every even column/.append style={column sep=0.2cm},draw=none},
			legend cell align={left},
		},
	}
	\newcommand{\figlabel}[2]{
		\node [anchor=north west,fill=white,inner sep=1pt,font=\scriptsize\strut] at ([xshift=0.5pt,yshift=-0.5pt]axis cs:#2) {#1};
	}

	\newcommand{\plotenv}[8]{
		\def\envpretty{#1}
		\def\env{#2}
		\def\ymax{#3}
		\def\pos{#4}
		\def\subfigurelabels{#5}
		\ifx&#6&
			\def\useylabel{ylabel={}}
		\else
			\def\useylabel{}
		\fi
		\ifx&#7&
			\def\thelegend{}
		\else
			\def\thelegend{\legend{Isolated,Homogeneous,Heterogeneous}}
		\fi
		\ifx&#8&
			\def\usexlabelfirstrow{}
		\else
			\def\usexlabelfirstrow{xlabel={}}
		\fi
		\def\xspacing{3.0cm}
		\def\yspacing{-2.5cm}
	
		\node [figtitle] at ($(\pos)+(1.4cm,0cm)$) {\Powerful};
		\node [figtitle] at ($(\pos)+(1.4cm+\xspacing,0cm)$) {\Weak};
		\node [figtitle] at ($(\pos)+(-0.1cm+\xspacing,0.35cm)$) {\envpretty};

		\begin{axis}[
				rlaxis,
				ymax=\ymax,
				at={(\pos)},
				\usexlabelfirstrow,
				\useylabel,
				]
			\ataricurve{isolated}{data/dqn_\env_large.csv}{dev1}
			\ataricurve{homogeneous}{data/dqn_\env_small3_small3.csv}{dev1}
			\ataricurve{heterogeneous}{data/dqn_\env_large_small3.csv}{dev1}

			\figlabel{(\StrMid{\subfigurelabels}{1}{1})}{0,\ymax}
		\end{axis}
		\begin{axis}[
				rlaxis,
				ymax=\ymax,
				at={($(\pos)+(\xspacing,0cm)$)},
				\usexlabelfirstrow,
				ylabel={},
				yticklabels={,,},
				]
			\ataricurve{isolated}{data/dqn_\env_small3.csv}{dev1}
			\ataricurve{homogeneous}{data/dqn_\env_small3_small3.csv}{dev2}
			\ataricurve{heterogeneous}{data/dqn_\env_large_small3.csv}{dev2}

			\thelegend

			\figlabel{(\StrMid{\subfigurelabels}{2}{2})}{0,\ymax}
		\end{axis}

		\ifx&#8&\else
			\begin{axis}[
					rlaxis,
					ymax=\ymax,
					at={($(\pos)+(0cm,\yspacing)$)},
					xlabel={},
					\useylabel,
					]
				\ataricurve{isolated}{data/dqn_\env_large.csv}{dev1}
				\ataricurve{homogeneous}{data/dqn_\env_small3_small3_slowdown_1_2.csv}{dev1}
				\ataricurve{heterogeneous}{data/dqn_\env_large_small3_slowdown_1_2.csv}{dev1}
	
				\figlabel{(\StrMid{\subfigurelabels}{3}{3}): $v_{weak}{=}0.5$}{0,\ymax}
			\end{axis}
			\begin{axis}[
					rlaxis,
					ymax=\ymax,
					at={($(\pos)+(\xspacing,\yspacing)$)},
					xmax=5,
					xlabel={},
					ylabel={},
					yticklabels={,,},
					]
				\ataricurve{isolated}{data/dqn_\env_small3.csv}{dev1}
				\ataricurve{homogeneous}{data/dqn_\env_small3_small3_slowdown_1_2.csv}{dev2}
				\ataricurve{heterogeneous}{data/dqn_\env_large_small3_slowdown_1_2.csv}{dev2}

				\figlabel{(\StrMid{\subfigurelabels}{4}{4}): $v_{weak}{=}0.5$}{0,\ymax}
			\end{axis}

			\begin{axis}[
					rlaxis,
					ymax=\ymax,
					at={($(\pos)+(0,2*\yspacing)$)},
					\useylabel,
					]
				\ataricurve{isolated}{data/dqn_\env_large.csv}{dev1}
				\ataricurve{blue}{data/dqn_\env_small3_small3_slowdown_1_4.csv}{dev1}
				\ataricurve{heterogeneous}{data/dqn_\env_large_small3_slowdown_1_4.csv}{dev1}

				\figlabel{(\StrMid{\subfigurelabels}{5}{5}): $v_{weak}{=}0.25$}{0,\ymax}
			\end{axis}
			\begin{axis}[
					rlaxis,
					ymax=\ymax,
					at={($(\pos)+(\xspacing,2*\yspacing)$)},
					xmax=2.5,
					ylabel={},
					yticklabels={,,},
					]
				\ataricurve{isolated}{data/dqn_\env_small3.csv}{dev1}
				\ataricurve{homogeneous}{data/dqn_\env_small3_small3_slowdown_1_4.csv}{dev2}
				\ataricurve{heterogeneous}{data/dqn_\env_large_small3_slowdown_1_4.csv}{dev2}

				\figlabel{(\StrMid{\subfigurelabels}{6}{6}): $v_{weak}{=}0.25$}{0,\ymax}
			\end{axis}
		\fi
	}
	\begin{tikzpicture}[
			figtitle/.style={
				anchor=south,inner sep=1pt,font=\footnotesize\strut\bf
			}
			]
		\plotenv{Breakout}{breakout}{170}{0,0}{abcdef}{ylabel}{legend}{slowdown}
		\plotenv{BeamRider}{beamrider}{2000}{6.9cm,0}{gh}{}{}{}
		\plotenv{SpaceInvaders}{spaceinvaders}{600}{6.9cm,-5.0cm}{ij}{}{}{}
	\end{tikzpicture}
	\caption{
		(a+b)~Heterogeneous learning achieves a high reward on both the \powerful and \weak devices on \emph{Atari 2600 Breakout}.
		With isolated learning, the \weak device suffers from instability.
		With homogeneous learning, the \powerful device suffers from low reward because it needs to employ a lightweight \ac{NN} and does not make use of its available resources.
		(c-f)~The \weak device interacts with its environment at a lower rate.
		Heterogeneous learning maintains a high reward on the \powerful device while enabling much faster convergence for the \weak device.
		(g-j)~Results on other games from the \emph{Atari 2600} suite show similar trends.
		\textbf{Key:}~Lines show the median test episode reward over 5~experiments with different random seeds.
		Shaded areas show minimum and maximum.
		Test episode rewards are measured every 10.000~steps and averaged with a sliding window of 25.
	}
	\label{fig:rl_hetero}
\end{figure}
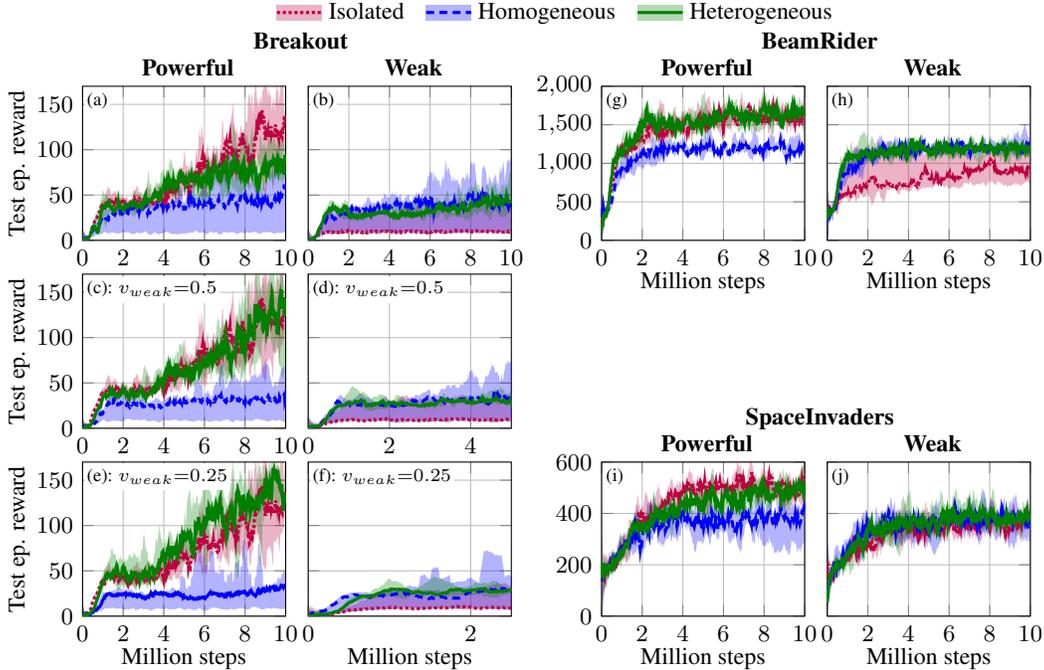

\cref{fig:rl_hetero}a and \ref{fig:rl_hetero}b present the results for \emph{Atari 2600 Breakout} when the \powerful and \weak devices interact with their environment at the same rate.
With isolated learning, the \powerful device achieves a higher test reward than the \weak device as it employs a complex \ac{NN}.
The \weak device even suffers from instability in the training, which causes most of the runs to not converge at all (reward does not improve).
A homogeneous setup considerably improves the convergence on the \weak device, as it can benefit from experiences made on the \powerful device.
However, as the \powerful device is restricted to also use a lightweight \ac{NN}, its resources are underutilized and its experienced reward significantly degrades.
Heterogeneous learning enables the \powerful device to employ a complex \ac{NN}, which improves its performance to almost the same level that was achieved in isolated execution while enabling the \weak device to reach a performance comparable to the homogeneous setup.
Thereby, heterogeneous learning achieves a high reward on \emph{both} devices at the same time, which neither isolated not homogeneous learning can obtain.
Most importantly, this is achieved by synchronizing \emph{only the first few layers} of the \acp{NN}, reducing its communication overhead compared with homogeneous learning. 
\cref{fig:rl_hetero}g-\ref{fig:rl_hetero}j show similar results for \emph{Atari 2600 BeamRider} and \emph{SpaceInvaders}.

\cref{fig:rl_hetero}c and \ref{fig:rl_hetero}d present results when the \weak device interacts with its environment at a rate of $v_{weak}{=}0.5{\times}$.
Therefore, it performs only 5~million interactions, while the \powerful device still performs 10~million interactions.
Homogeneous and heterogeneous learning can deal with this heterogeneity, enabling the \weak device to converge much faster than with isolated learning.
Importantly, however, for heterogeneous learning, this does not come at a lower reward on the \powerful device.
We study in \cref{fig:rl_hetero}c and \ref{fig:rl_hetero}d an even lower rate of interaction ($0.25{\times}$) on the \weak device, and observe even larger improvements over isolated learning.
In summary, heterogeneous learning allows us to get the best of both worlds, outperforming the baselines in all different scenarios.

\subsection{Supervised learning}
\label{sec:sl}

In this section, we evaluate the performance of heterogeneous learning on an image classification task using supervised learning.  
We select the \emph{CIFAR-10} dataset~\cite{cifar10} for demonstration, containing 50,000~images of 10~classes for training, and 10,000~images for testing.
We employ a commonly-used state-of-the-art solution and study how heterogeneous learning affects the achievable accuracy, compared with the baselines.

\pgfplotsset{
	sl_final_axis/.style={
		anchor=north west,
		width=2.8cm,
		height=2.2cm,
		xmin=0.05*\xmax,
		xmax=0.5*\xmax,
		ymin=56,
		ymax=84,
		scaled ticks=false,
		xtick={10,20,30,40,50},
		xlabel={Amt. of train. data \axisunit{\%}},
		xlabel style={align=center},
		ylabel={Test accuracy [\%]},
		legend style={
			at={(1.0,1.0)},
			anchor=north west,
			legend columns=1,
			xshift=0.3cm,
			yshift=0.0cm,
		},
		legend style={/tikz/every even column/.append style={column sep=0.2cm},draw=none},
		legend cell align={left},
		xmajorgrids,
		ymajorgrids,
	},
	sl_learning_axis/.style={
		anchor=north west,
		width=2.8cm,
		height=2.2cm,
		xmin=0,
		xmax=4,
		ymin=56,
		ymax=84,
		scaled ticks=false,
		xtick distance=1,
		xlabel={Thousand rounds},
		xlabel style={align=center},
		ylabel style={align=center},
		legend style={
			at={(1.0,1.0)},
			anchor=south,
			legend columns=3,
			xshift=0.1cm,
		},
		legend style={/tikz/every even column/.append style={column sep=0.2cm},draw=none},
		legend cell align={left},
		xmajorgrids,
		ymajorgrids,
	}
}
\newcommand{\figlabel}[2]{
	\node [anchor=north west,fill=white,inner sep=1pt,font=\footnotesize\strut] at ([xshift=0.5pt,yshift=-0.5pt]axis cs:#2) {#1};
}

\begin{figure}
	\centering
	\newcommand{\finalaccuracy}[9]{
		\def\pos{#1}
		\def\thetitle{#2}
		\def\powerfulmodel{#3}
		\def\weakmodel{#4}
		\def\merging{#5}
		\def\subfigurelabels{#6}
		\ifx&#7&
			\def\usexlabel{xlabel={}}
		\else
			\def\usexlabel{}
		\fi
		\ifx&#8&
			\def\useylabel{ylabel={}}
		\else
			\def\useylabel{}
		\fi
		\ifx&#9&
			\def\thelegend{}
		\else
			\def\thelegend{\legend{Isolated,Homogeneous,Heterogeneous}}
		\fi

		\def\xspacing{3.15cm}

		\node [figtitle] at ($(\pos)+(-0.1cm+\xspacing,0.0cm)$) {\thetitle};

		\begin{axis}[
				sl_final_axis,
				at={(\pos)},
				xticklabels={90,80,70,60,50},
				\usexlabel,
				\useylabel,
				]
			\drawstd{isolated}{data/supervised_\powerfulmodel_\weakmodel_isolated.csv}{model1_accuracy}

			\drawstd{homogeneous}{data/supervised_\weakmodel_\weakmodel_\merging_merging.csv}{model1_accuracy}

			\drawstd{heterogeneous}{data/supervised_\powerfulmodel_\weakmodel_\merging_merging.csv}{model1_accuracy}

			\figlabel{(\StrMid{\subfigurelabels}{1}{1}) \Powerful}{5,84}
		\end{axis}
		\begin{axis}[
				sl_final_axis,
				at={($(\pos)+(\xspacing,0cm)$)},
				\usexlabel,
				ylabel={},
				yticklabels={,,},
				]
			\drawstd{isolated}{data/supervised_\powerfulmodel_\weakmodel_isolated.csv}{model2_accuracy}

			\drawstd{homogeneous}{data/supervised_\weakmodel_\weakmodel_\merging_merging.csv}{model2_accuracy}

			\drawstd{heterogeneous}{data/supervised_\powerfulmodel_\weakmodel_\merging_merging.csv}{model2_accuracy}

			\figlabel{(\StrMid{\subfigurelabels}{2}{2}) \Weak}{5,84}

			\thelegend
		\end{axis}
	}

	\NewDocumentCommand{\sllearning}{ommmm}{
		\def\thetitle{#2}
		\def\frac{#3}
		\def\pos{#4}
		\def\subfigurelabels{#5}

		\def\xspacing{3.15cm}

		\node [figtitle] at ($(\pos)+(-0.1cm+\xspacing,0.0cm)$) {\thetitle};

		\begin{axis}[
				sl_learning_axis,
				at={(\pos)},
				ylabel={},
				\IfNoValueTF{#1}{yticklabels={,,},}{}
				]
			\supervisedlearningcurve{isolated}{data/supervised_iterations_c-large3-d2_c-tiny_\frac_isolated.csv}{model1_test_accuracy}

			\supervisedlearningcurve{homogeneous}{data/supervised_iterations_c-tiny_c-tiny_\frac_static_merging.csv}{model1_test_accuracy}

			\supervisedlearningcurve{heterogeneous}{data/supervised_iterations_c-large3-d2_c-tiny_\frac_static_merging.csv}{model1_test_accuracy}

			\figlabel{(\StrMid{\subfigurelabels}{1}{1}) \Powerful}{0,84}
		\end{axis}
		\begin{axis}[
				sl_learning_axis,
				at={($(\pos)+(\xspacing,0cm)$)},
				ylabel={},
				yticklabels={,,}
				]
			\supervisedlearningcurve{isolated}{data/supervised_iterations_c-large3-d2_c-tiny_\frac_isolated.csv}{model2_test_accuracy}

			\supervisedlearningcurve{homogeneous}{data/supervised_iterations_c-tiny_c-tiny_\frac_static_merging.csv}{model2_test_accuracy}

			\supervisedlearningcurve{heterogeneous}{data/supervised_iterations_c-large3-d2_c-tiny_\frac_static_merging.csv}{model2_test_accuracy}

			\figlabel{(\StrMid{\subfigurelabels}{2}{2}) \Weak}{0,84}
		\end{axis}
	}

	\begin{tikzpicture}[
			figtitle/.style={
				anchor=south,fill=white,inner sep=1pt,font=\footnotesize\strut\bf
			}
			]
		\finalaccuracy{0cm,0cm}{Share only first layers}{large3}{tiny}{static}{ab}{}{ylabel}{legend}
		\finalaccuracy{0cm,-3.1cm}{Cascading}{c-large3-d2}{c-tiny}{static}{cd}{xlabel}{ylabel}{}
		\sllearning[yticks]{Cascading training progress (data: 80\,\% / 20\,\%)}{0.20}{6.6cm,-3.1cm}{ef}
	\end{tikzpicture}
	\caption{
		(a+b) Heterogeneous learning improves the accuracy on the \powerful device over a homogeneous setup at the cost of accuracy loss on the \weak device.
		(c+d) Heterogeneous learning with a cascaded topology outperforms both homogeneous and isolated learning.
		(e+f) Heterogeneous learning achieves a fast convergence on both devices.
		\textbf{Key:}~Lines show the median accuracy over 5 repetitions with different random seeds, shaded areas show minimum and maximum.
	}
	\label{fig:supervised_hetero}
\end{figure}
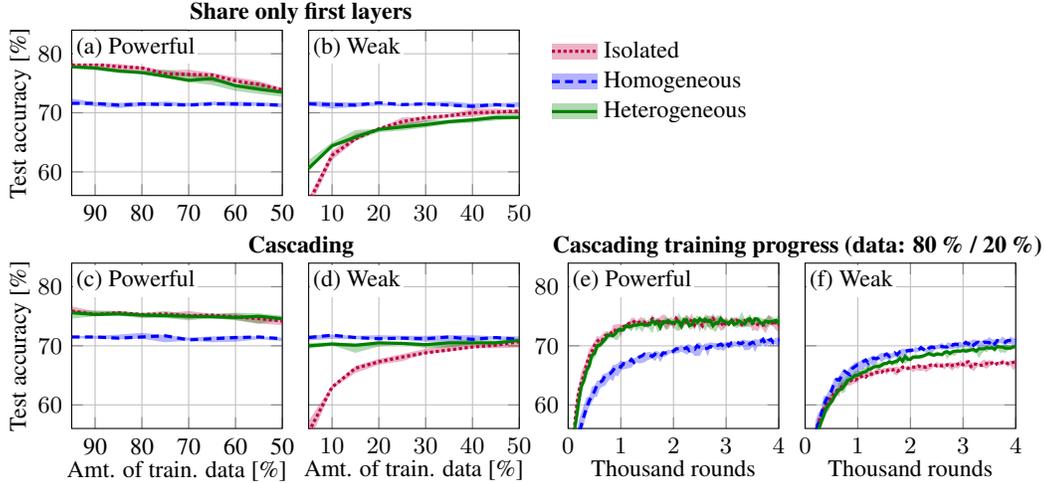

Similarly to the previous section, we consider two types of devices with different resource profiles, namely \emph\powerful~and \emph\weak, and create two different \ac{NN} topologies (\emph{complex} and \emph{lightweight}) with different numbers of parameters (memory requirements) and operations (computational/energy/etc.\ requirements) that are modeled to fit the available resources of the respective devices.
To model heterogeneous availability of training data, we divide the training data into two disjoint sets, distributed among the devices.
Each device then splits its local data into 80\,\% for training and 20\,\% for validation.
During each round, each device trains its local \ac{NN} with a batch of 2,000~randomly-sampled examples from its local training data.
Then, all devices synchronize their shared parameters with the help of a synchronous coordinator.
For homogeneous learning, we employ the state-of-the-art \emph{FedAvg} algorithm~\cite{federated}, employed by the coordinator, which uses merging weights proportional to the amount of data on each device to avoid bias towards few training examples.
Similarly, we have adjusted the merging mechanism for heterogeneous learning, using a weighted summation to merge the shared parameters, as detailed in the supplementary materials.
We run the system for 4,000~rounds.
After each round, each device independently validates its local \ac{NN} based on local validation data to select the best version of the local \ac{NN}.
After training, we evaluate all \acp{NN} with the test data of \emph{CIFAR-10}.

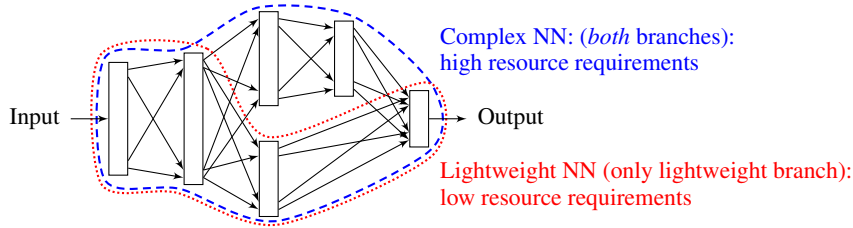
\begin{figure}
	\centering
	\begin{tikzpicture}
		\node[nnlayer,minimum height=1.5cm] (l1) at (0,0) {};
		\node[nnlayer,minimum height=1.75cm] (l2) at (1,0) {};
		\node[nnlayer,minimum height=1.25cm] (l31) at (2,0.8) {};
		\node[nnlayer,minimum height=1.0cm] (l41) at (3,0.8) {};

		\node[nnlayer,minimum height=1cm] (l32) at (2,-0.8) {};

		\node[nnlayer,minimum height=0.75cm] (out) at (4,0.0) {};

		\nnconnect{l1}{l2}
		\nnconnect{l2}{l31}
		\nnconnect{l2}{l32}
		\nnconnect{l31}{l41}
		\nnconnect{l41}{out}
		\nnconnect{l32}{out}

		\draw [blue,thick,densely dashed] plot [smooth cycle,tension=0.5] coordinates {
			([xshift=-0.5mm,yshift=0.5mm]l1.north west)
			([yshift=0.5mm]l2.north west)
			([yshift=0.5mm]l31.north west)
			([yshift=0.5mm]l41.north east)
			([xshift=0.5mm,yshift=0.5mm]out.north east)
			([yshift=-0.5mm]out.south east)
			([yshift=-0.5mm]l32.south east)
			([xshift=-0.5mm,yshift=-0.5mm]l2.south west)
			([xshift=-0.5mm,yshift=-0.5mm]l1.south west)
		};
		\draw [red,thick,densely dotted] plot [smooth cycle,tension=0.5] coordinates {
			([xshift=-1mm,yshift=1mm]l1.north west)
			([xshift=0.5mm,yshift=0.5mm]l2.north east)
			([yshift=0.5mm]l32.north)
			([xshift=1mm,yshift=1mm]out.north)
			([xshift=0.5mm,yshift=-1mm]out.south east)
			([yshift=-1mm]l32.south east)
			([xshift=-1mm,yshift=-1mm]l2.south west)
			([xshift=-1mm,yshift=-1mm]l1.south west)
		};

		\draw[latex'-] (l1.west) to ++(-5mm,0) node[label,left] {Input};
		\draw[-latex'] (out.east) to ++(5mm,0) node[label,right] {Output};

		\node [label,blue,align=left,anchor=south west] at ([yshift=0.5mm]out.north east) {Complex NN: (\emph{both} branches):\\high resource requirements};
		\node [label,red,align=left,anchor=north west] at ([yshift=-0.5mm]out.south east) {Lightweight NN (only lightweight branch):\\low resource requirements};
	\end{tikzpicture}
	\caption{
		A cascaded topology includes the lightweight branch into the complex NN to allow sharing of all parameters of the lightweight branch in heterogeneous learning.
	}
	\label{fig:cascading}
\end{figure}

We first employ the same technique for parameter sharing in heterogeneous learning as for \ac{RL}, i.e., share only the first few layers.
We base the complex topology and the training hyperparameters on~\cite{cifar10_large}.
The lightweight topology shares the first 6 layers with the complex network but then has fewer and smaller following layers.
Thereby, the number of computational operations and parameters of the lightweight \ac{NN} are reduced to 57.2\,\% and 6.7\,\% of the complex \ac{NN}, respectively.
\cref{fig:supervised_hetero}a and~\ref{fig:supervised_hetero}b present the results.
We sweep the fraction of training data present on the \weak device from 5\,\% to 50\,\%, the remainder of the data is available on the \powerful device.
We observe that homogeneous learning achieves a constant accuracy irrespective of how the training data is distributed.
Heterogeneous learning improves the accuracy on the \powerful device over homogeneous learning because it employs a complex \ac{NN}.
However, these benefits come at the cost of a lower accuracy on the \weak device.
The reason is that heterogeneous learning shares and synchronizes fewer parameters than homogeneous learning.
Especially, not sharing the output layers causes these layers to overfit if little training data is available on the weak device.
To overcome this limitation, we adjust the way the two \acp{NN} share layers, as depicted in \cref{fig:cascading}.
The lightweight \ac{NN} is fully-contained in the complex \ac{NN}, which now has two branches and gets slightly more complex.
Therefore, the lightweight \ac{NN} now has 57.1\,\% of the computational operations and 6.4\,\% of the parameters compared to the complex \ac{NN}.
All topologies are listed in detail in the supplementary material.
With this cascaded topology, heterogeneous learning can synchronize all parameters of the lightweight \ac{NN}.
This topology is similar to the grid modules in \emph{Inception-ResNet}~\cite{inception_resnet} but applied on a larger scale.
We apply additional dropout~\cite{dropout} at the end of the complex branch that drops the whole branch with a probability of 0.5.
Thereby, co-adaptations, in which the lightweight branch depends on the output of the complex branch, are avoided as the lightweight branch needs to be able to work on its own.

\cref{fig:supervised_hetero}c and \ref{fig:supervised_hetero}d show the results with this cascaded topology and the data-dependent parameter merging.
We observe that on the \powerful device, isolated learning and heterogeneous learning both outperform homogeneous learning because they employ the complex \ac{NN} instead of the lightweight \ac{NN}.
On the \weak device, homogeneous and heterogeneous learning both outperform isolated learning because they enable to share parameters with the \powerful device enabling to benefit from its training data.
This advantage gets more important the less data is available on the \weak device.
Heterogeneous learning always achieves a high accuracy and, therefore, is preferable.
\cref{fig:supervised_hetero}f and \ref{fig:supervised_hetero}g show how the test accuracy evolves during training for a data distribution of 80\,\% on the \powerful device and 20\,\% on the \weak device.
Heterogeneous learning achieves a fast convergence on both devices.

In summary, careful engineering is required when adapting our solution to supervised learning tasks.
Specifically, we observe that the benefits of heterogeneous learning strongly depend on how parameters are shared among devices and merged.
Such observations have not been made for \ac{RL}.
A fundamental difference between the two is that in supervised learning, the training data is limited, whereas in \ac{RL} the rate at which new data is gathered is limited.
Overfitting may occur with supervised learning when training is performed using the same set of training data over and over.
\ac{RL} does not suffer from such a problem, partly because it continuously generates new training data over, preventing biases toward a specific subset of data~\cite{sutton1998introduction}. 
This could to some extent explain the differences we observed when applying our solution to these two different tasks.
These complexities aside, the relative gain is similar, strongly supporting our solution.

\section{Conclusion}
\label{sec:conclusion}

We introduced cooperative learning tailored for heterogeneous distributed systems, where each device trains an \ac{NN} which fits its resource profile and constraints, and where the cooperation among different devices is made possible by sharing some layers among all \acp{NN}.
We have studied the performance of our solutions on different \ac{RL} and supervise learning tasks, showing that it significantly outperforms state-of-the-art solutions, as all devices, low-end or high-end, can experience a better reward/accuracy.
Our solution extends the application of distributed learning toward end-users, where the data is generated, localizing the training and enabling new use cases.

\clearpage
\section*{Broader impact}

Heterogeneous distributed learning allows low-end devices to benefit from what is learned in more powerful devices, but unlike the homogeneous learning, this does not come with a performance cost at powerful devices.
Hence, it enables the inclusion of low-end, and thus low-cost, devices in a cooperative learning system, without any specific overhead, if we ignore the related communication overheads for exchanging the shared parameters, for the other members of the system. 
Our solution enables many new use cases.
For instance, this allows smartphone users that cannot afford high-end devices (e.g., due to a low income) to benefit from better experiences, by letting them learn from experiences of users with high-end smartphones.
As another example, distributed sensor networks may be implemented with a mixture of low-cost and more powerful devices, lowering the overall costs, which may for instance enable more fine-grained environmental monitoring.

The impact of distributed heterogeneous learning on the carbon footprint of our society is ambivalent.
Training \acp{NN} is a very energy-consuming task.
On the one hand, distributing the training over low-end devices, which are not particularly designed and optimized for such computations, could increase the overall energy consumption of the system for the training.
On the other hand, however, localizing the training reduces the communication overhead, and hence, related energy costs of transmitting locally gathered data to other devices or to a centralized data center.
There is clearly a trade-off between these two, which should be studied in detail, shall such system be deployed on a large scale.
Besides, according to Jevens paradox, enabling the inclusion of more devices likely increases the demand, i.e., increases the usage of \acp{NN}, which increases the carbon footprint.

Furthermore, distributed learning imposes the risk of malicious users sabotaging or undermining the learned models.
In a straightforward way, malicious users may send wrong parameter updates to the coordinator or other devices -- harming the performance of all other devices that participate in distributed learning.
In a more sophisticated attack, malicious users may include hard-to-detect backdoors into the jointly-learned model by training the model to be susceptible to certain off-distribution patterns -- enabling the attack to devices that participate in the system.
For instance, an \ac{NN} that detects and classifies traffic sign detection can be trained to misclassify traffic signs with a yellow post-it~\cite{badnets}, which can be exploited by attackers to disturb the traffic or even cause accidents. This problem is not however specific to our solution and is related to distributed learning in general.

\begin{ack}
Martin Rapp and Ramin Khalili are corresponding authors.
This work is in parts funded by the Deutsches Bundesministerium für Bildung und Forschung (BMBF, Federal Ministry of Education and Research in Germany) and the Deutsche Forschungsgemeinschaft (DFG, German Research Foundation) -- Project Number 146371743 -- TRR 89 Invasive Computing.
\end{ack}

\bibliographystyle{IEEEtran}
{\small \bibliography{bibliography}}
\appendix

\renewcommand{\floatpagefraction}{.8}%
\section{Heterogeneous availability of training data in supervised learning}
\label{sec:hetero_data}

In the case of supervised learning, a different amount of training data may be available on each device.
Due to this unbalanced distribution of training data, we need to modify how the \ac{NN} parameters are merged.
This observation is not new and, e.g., the state-of-the-art \emph{FedAvg}~\cite{federated} algorithm takes the amount of data at each device into account.
Since we modify the way how parameters are shared between devices compared to homogeneous learning techniques, we briefly revisit this here.

In supervised learning, each device~$k$ calculates the gradient~$d\theta^{(k)}$ of the parameters~$\theta$ w.r.t.\ the loss function $L$ on its own training data~$D^{(k)}$, which is a subset of all training data~$D$ of all devices.
\begin{equation}
	D = \bigcup_{k=1}^{n} D^{(k)}
\end{equation}
Vanilla gradient descent is rarely used, but variants of it that estimate the gradient using a subset (batch) of the training data.
However, the expected value of the gradient still equals the true gradient, i.e.
\begin{equation}
	E(d\theta^{(k)}) = \frac{1}{|D^{(k)}|} \sum_{d \in D^{(k)}} \frac{\partial}{\partial\theta} L(d;\theta).
\end{equation}
During synchronization, parameter updates~$d\theta^{(k)}$ from all devices are combined.
If devices have different amount of data $|D^{(k)}|$, the updates from all $n$~devices must be combined using weighted average with weights~$\alpha^{(k)}$, which are proportional to the amount of data on the $k$-th device $|D^{(k)}|$
This is done to avoid bias towards training examples on a device with little training data:
\begin{align}
	\alpha^{(k)} &= \frac{|D^{(k)}|}{|D|}
	\\
	d\theta
	&= \sum_{k=1}^{n}\left( \alpha^{(k)} \cdot d\theta^{(k)}\right)
	= \sum_{k=1}^{n}\left( \frac{|D^{(k)}|}{|D|} \cdot d\theta^{(k)}\right),
\end{align}
where $n$ is the number of devices, hence,
\begin{align}
	E(d\theta) &= \sum_{k=1}^{n}\left( \frac{|D^{(k)}|}{|D|} \cdot E\left(d\theta^{(k)}\right)\right)\\ 
	&= \sum_{k=1}^{n}\left(\frac{|D^{(k)}|}{|D|} \cdot \frac{1}{|D^{(k)}|} \sum_{d \in D^{(k)}} \frac{\partial}{\partial\theta} L(d;\theta)\right) \\
	&= \frac{1}{|D|} \sum_{k=1}^{n}\sum_{d \in D^{(k)}} \frac{\partial}{\partial\theta} L(d;\theta) \\
	&= \frac{1}{|D|} \sum_{d \in D} \frac{\partial}{\partial\theta} L(d;\theta)
\end{align}
Thereby, distributed training behaves similar to a global gradient descent.
This is only achieved if merging weights~$\alpha^{(k)}$ are proportional to the amount of data on each device.

\section{Implementation details}

This section lists implementation details such as \ac{NN} topologies and hyperparameters used in the experiments section.

\subsection{Reinforcement learning details}

We use the \emph{Atari-2600} games~\cite{atari} as a benchmark.
The observable state is the screen output of the game, which is a $84{\times}84$ pixels gray-scale image.
The actions of the agent are the buttons to press, where only one button is pressed at a time.
We repeat the same action for four consecutive frames, and also use a concatenation of the last four frames as input for the \ac{NN}, i.e., the \ac{NN} input size is $84{\times}84{\times}4$.
We scale the pixel values to the range~$[0,1]$.

\begin{figure}
	\centering
	\def\step{0.3cm}
	\begin{tikzpicture}[
			layer/.style={draw,inner sep=0pt, outer sep=0pt,minimum width=3.7cm, minimum height=0.45cm,font=\footnotesize\strut},
			shared/.style={},
		]
		\node[layer,shared] (sh1) {Conv $8{\times}8\times32, \text{stride}{=}4$};
		\node[layer,shared,below=\step of sh1.south,anchor=north] (sh2) {ReLU};
		\node[layer,shared,below=\step of sh2.south,anchor=north] (sh3) {Conv $4{\times}4\times64, \text{stride}{=}2$};
		\node[layer,shared,below=\step of sh3.south,anchor=north] (sh4) {ReLU};

		\node[layer,below left=\step and 2.2cm of sh4.south,anchor=north] (l1) {Conv $3{\times}3\times64, \text{stride}{=}1$};
		\node[layer,below=\step of l1.south,anchor=north]  (l2) {ReLU};
		\node[layer,below=\step of l2.south,anchor=north]  (l3) {Dense $512$};
		\node[layer,below=\step of l3.south,anchor=north]  (l4) {ReLU};
		\node[layer,below=\step of l4.south,anchor=north]  (l5) {Dense $6$};

		\draw[latex'-] (sh1.north) to ++(0,\step) node [above,inner sep=2pt,font=\footnotesize\strut] {Input: $84\times84\times4$};
		\draw[-latex'] (sh1)  to (sh2);
		\draw[-latex'] (sh2)  to (sh3);
		\draw[-latex'] (sh3)  to (sh4);
		\draw[-latex'] (sh4.south) -- ++(0,-0.3*\step) -| (l1.north);
		\draw[-latex'] (l1)   to (l2);
		\draw[-latex'] (l2)   to (l3);
		\draw[-latex'] (l3)   to (l4);
		\draw[-latex'] (l4)   to (l5);
		\draw[-latex'] (l5.south) to ++(0,-\step) node [right,anchor=north,font=\footnotesize\strut,inner sep=2pt,align=center] {Complex NN output\\[2pt]Parameters: 1,687,206\\Operations: 9,389,318};

		\node[layer,below right=\step and 2.2cm of sh4.south,anchor=north] (t1) {Conv $3{\times}3\times8, \text{stride}{=}1$};
		\node[layer,below=\step of t1.south,anchor=north]  (t2) {ReLU};
		\node[layer,below=\step of t2.south,anchor=north]   (t3) {Dense $64$};
		\node[layer,below=\step of t3.south,anchor=north]   (t4) {ReLU};
		\node[layer,below=\step of t4.south,anchor=north]   (t5) {Dense $6$};

		\draw[-latex'] (sh4.south) -- ++(0,-0.3*\step) -| (t1.north);
		\draw[-latex'] (t1)   to (t2);
		\draw[-latex'] (t2)   to (t3);
		\draw[-latex'] (t3)   to (t4);
		\draw[-latex'] (t5.south) to ++(0,-\step) node [right,anchor=north,font=\footnotesize\strut,inner sep=2pt,align=center] {Lightweight NN output\\[2pt]Parameters: 71,214\\Operations: 6,219,158};
	\end{tikzpicture}
	\caption{Topologies of the \acp{NN} for the \emph{Atari-2600} games.}
	\label{fig:atari_topo}
\end{figure}
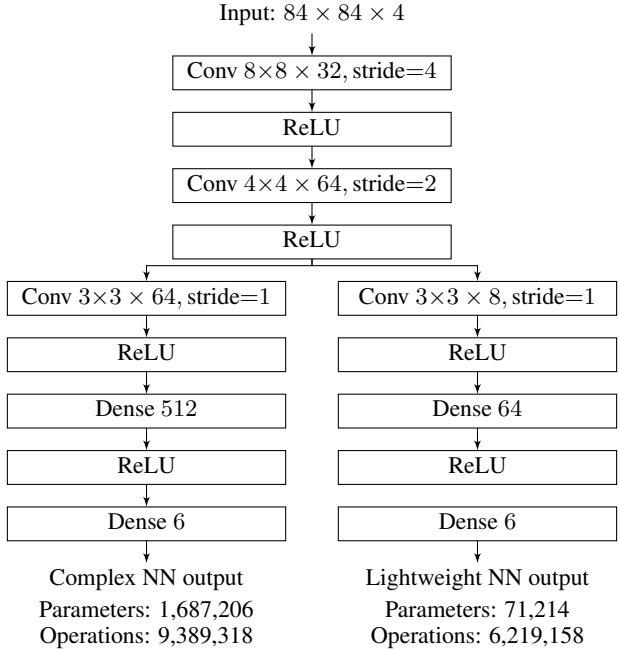

\cref{fig:atari_topo} presents the topologies of the complex and lightweight \acp{NN}.
The complex \ac{NN} is based on~\cite{rl_atari}.
The lightweight \ac{NN} shares the first four layers.
The following layers have a reduced number of channels / neurons.
Thereby, the number of parameters and computational operations is reduced to 66\,\% and 4.2\,\%, respectively.
We use the Adam~\cite{adam} optimizer and set the learning rate at $\eta=0.00025$.
We employ Huber loss, with $\delta = 1$.
\begin{equation}
	L_\delta(y_{true},y_{pred})=\begin{cases}
	0.5 \cdot (y_{true}-y_{pred})^2 &|y_{true}-y_{pred}| \le \delta,\\
	\delta \cdot |y_{true}-y_{pred}| - 0.5 \cdot \delta &\text{otherwise}.
	\end{cases}
\end{equation}

We base the training and \ac{RL} hyperparameters also on~\cite{rl_atari}.
We employ \ac{DDQL} with experience replay.
The size of the reply memory is 1,000,000 steps for the complex \ac{NN}, and 100,000 steps for the lightweight \ac{NN}.
We do not perform any training during a warmup phase of 50,000 steps.
The batch size for a single step of training is 32.
The updated parameters are copied to the target network every 10,000 steps.
We set the \ac{RL} discount factor $\gamma=0.99$.
We use $\epsilon$-greedy policy with decaying $\epsilon$.
$\epsilon$ starts at a value of 1.0 and decreases linearly to 0.1 within the first 1,000,000 steps, after which it remains at 0.1.
For testing, we temporally reduce $\epsilon$ to~0.02.

We base our implementation on the \emph{keras-rl} library~\cite{kerasrl}, \emph{keras}~\cite{keras} and \emph{OpenAI~gym}~\cite{gym}.

\subsection{Supervised learning details}
\label{sec:app:cifar10}

We use the CIFAR-10~\cite{cifar10} dataset as a benchmark.
It contains colored images with $32{\times}32$ pixels from ten classes: airplanes, cars, birds, cats, deer, dogs, frogs, horses, ships, and trucks.
CIFAR-10 has 50,000 training/validation examples and 10,000 test examples.
We distribute the training/validation examples among devices.
Each device then splits its local data into 80\,\% for training and 20\,\% for validation.
The test data is not provided to the devices but used offline to evaluate the accuracy of model snapshots.
We scale the pixel values to the range $[0,1]$.
No further preprocessing is performed.
We report the accuracy (i.e., TOP-1 accuracy), which is the fraction of images where the highest probability was assigned to the correct class.

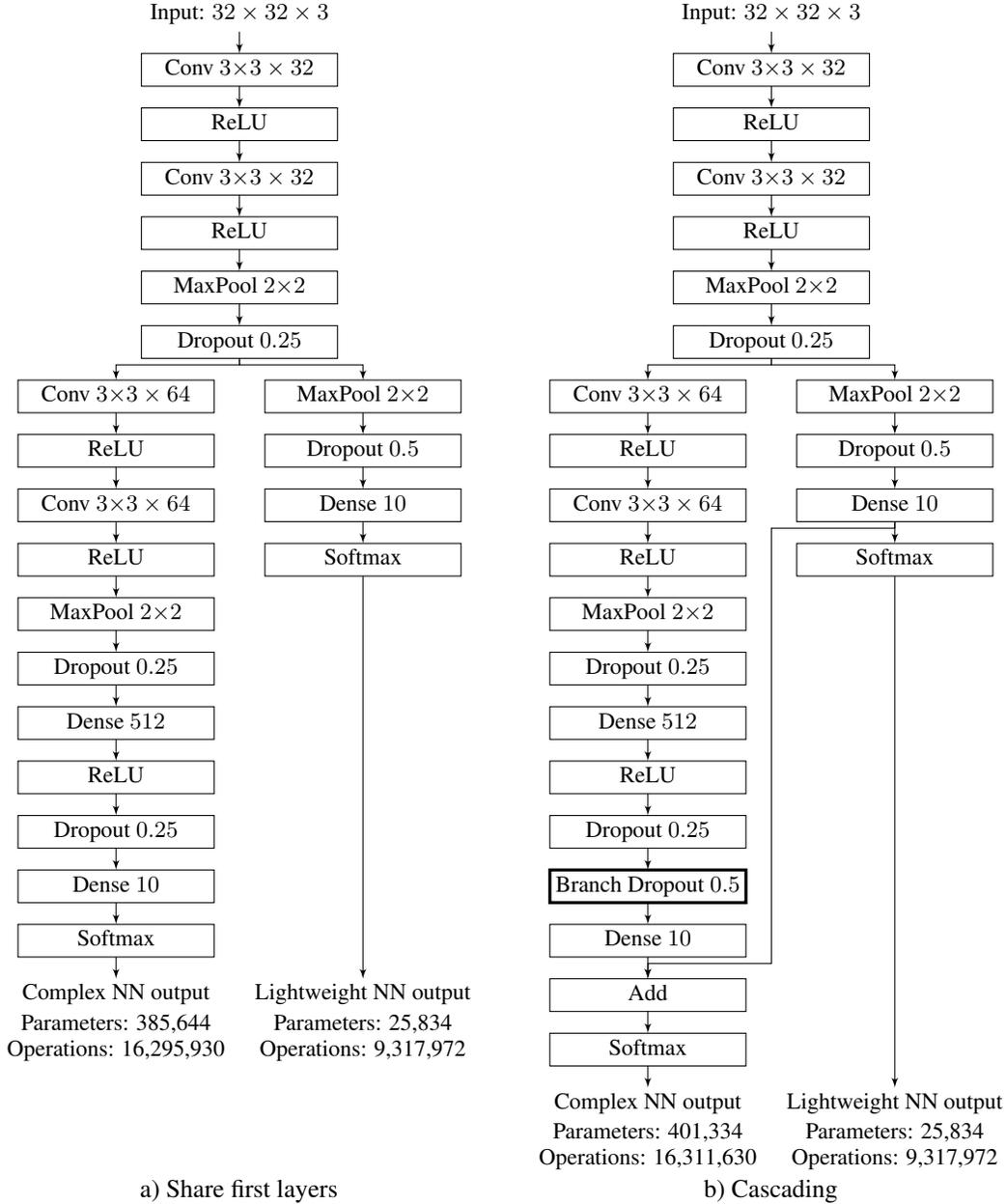
\begin{figure}
	\centering
	\def\step{0.3cm}
	\begin{tikzpicture}
		\node[container] (sharefirst) {
			\begin{tikzpicture}[
					layer/.style={draw,inner sep=0pt, outer sep=0pt,minimum width=2.7cm, minimum height=0.45cm,font=\footnotesize\strut},
					shared/.style={},
				]
				\node[layer,shared] (sh1) {Conv $3{\times}3\times32$};
				\node[layer,shared,below=\step of sh1.south,anchor=north] (sh2) {ReLU};
				\node[layer,shared,below=\step of sh2.south,anchor=north] (sh3) {Conv $3{\times}3\times32$};
				\node[layer,shared,below=\step of sh3.south,anchor=north] (sh4) {ReLU};
				\node[layer,shared,below=\step of sh4.south,anchor=north] (sh5) {MaxPool $2{\times}2$};
				\node[layer,shared,below=\step of sh5.south,anchor=north] (sh6) {Dropout $0.25$};
		
				\node[layer,below left=\step and 1.7cm of sh6.south,anchor=north] (l1) {Conv $3{\times}3\times64$};
				\node[layer,below=\step of l1.south,anchor=north]  (l2) {ReLU};
				\node[layer,below=\step of l2.south,anchor=north]  (l3) {Conv $3{\times}3\times64$};
				\node[layer,below=\step of l3.south,anchor=north]  (l4) {ReLU};
				\node[layer,below=\step of l4.south,anchor=north]  (l5) {MaxPool $2{\times}2$};
				\node[layer,below=\step of l5.south,anchor=north]  (l6) {Dropout $0.25$};
				\node[layer,below=\step of l6.south,anchor=north]  (l7) {Dense $512$};
				\node[layer,below=\step of l7.south,anchor=north]  (l8) {ReLU};
				\node[layer,below=\step of l8.south,anchor=north]  (l9) {Dropout $0.25$};
				\node[layer,below=\step of l9.south,anchor=north]  (l10) {Dense $10$};
				\node[layer,below=\step of l10.south,anchor=north] (l11) {Softmax};
		
				\draw[latex'-] (sh1.north) to ++(0,\step) node [above,inner sep=2pt,font=\footnotesize\strut] {Input: $32\times32\times3$};
				\draw[-latex'] (sh1)  to (sh2);
				\draw[-latex'] (sh2)  to (sh3);
				\draw[-latex'] (sh3)  to (sh4);
				\draw[-latex'] (sh4)  to (sh5);
				\draw[-latex'] (sh5)  to (sh6);
				\draw[-latex'] (sh6.south) -- ++(0,-0.3*\step) -| (l1.north);
				\draw[-latex'] (l1)   to (l2);
				\draw[-latex'] (l2)   to (l3);
				\draw[-latex'] (l3)   to (l4);
				\draw[-latex'] (l4)   to (l5);
				\draw[-latex'] (l5)   to (l6);
				\draw[-latex'] (l6)   to (l7);
				\draw[-latex'] (l7)   to (l8);
				\draw[-latex'] (l8)   to (l9);
				\draw[-latex'] (l9)   to (l10);
				\draw[-latex'] (l10)  to (l11);
				\draw[-latex'] (l11.south) to ++(0,-\step) node [right,anchor=north,font=\footnotesize\strut,inner sep=2pt,align=center] {Complex NN output\\[2pt]Parameters: 385,644\\Operations: 16,295,930};
		
				\node[layer,below right=\step and 1.7cm of sh6.south,anchor=north] (t1) {MaxPool $2{\times}2$};
				\node[layer,below=\step of t1.south,anchor=north]  (t2) {Dropout $0.5$};
				\node[layer,below=\step of t2.south,anchor=north]   (t3) {Dense $10$};
				\node[layer,below=\step of t3.south,anchor=north]   (t4) {Softmax};
		
				\draw[-latex'] (sh6.south) -- ++(0,-0.3*\step) -| (t1.north);
				\draw[-latex'] (t1)   to (t2);
				\draw[-latex'] (t2)   to (t3);
				\draw[-latex'] (t3)   to (t4);
				\draw[-latex'] (t4.south) to (t4.south |- l11.south) to ++(0,-\step) node [right,anchor=north,font=\footnotesize\strut,inner sep=2pt,align=center] {Lightweight NN output\\[2pt]Parameters: 25,834\\Operations: 9,317,972};
			\end{tikzpicture}
		};
		\node[container,right=0.8cm of sharefirst.north east,anchor=north west] (cascading) {
			\begin{tikzpicture}[
					layer/.style={draw,inner sep=0pt, outer sep=0pt,minimum width=2.7cm, minimum height=0.45cm,font=\footnotesize\strut},
					shared/.style={},
				]
				\node[layer,shared] (sh1) {Conv $3{\times}3\times32$};
				\node[layer,shared,below=\step of sh1.south,anchor=north] (sh2) {ReLU};
				\node[layer,shared,below=\step of sh2.south,anchor=north] (sh3) {Conv $3{\times}3\times32$};
				\node[layer,shared,below=\step of sh3.south,anchor=north] (sh4) {ReLU};
				\node[layer,shared,below=\step of sh4.south,anchor=north] (sh5) {MaxPool $2{\times}2$};
				\node[layer,shared,below=\step of sh5.south,anchor=north] (sh6) {Dropout $0.25$};

				\node[layer,below left=\step and 1.7cm of sh6.south,anchor=north] (l1) {Conv $3{\times}3\times64$};
				\node[layer,below=\step of l1.south,anchor=north]  (l2) {ReLU};
				\node[layer,below=\step of l2.south,anchor=north]  (l3) {Conv $3{\times}3\times64$};
				\node[layer,below=\step of l3.south,anchor=north]  (l4) {ReLU};
				\node[layer,below=\step of l4.south,anchor=north]  (l5) {MaxPool $2{\times}2$};
				\node[layer,below=\step of l5.south,anchor=north]  (l6) {Dropout $0.25$};
				\node[layer,below=\step of l6.south,anchor=north]  (l7) {Dense $512$};
				\node[layer,below=\step of l7.south,anchor=north]  (l8) {ReLU};
				\node[layer,below=\step of l8.south,anchor=north]  (l9) {Dropout $0.25$};
				\node[layer,below=\step of l9.south,anchor=north,very thick]  (l10) {Branch Dropout $0.5$};
				\node[layer,below=\step of l10.south,anchor=north]  (l11) {Dense $10$};
				\node[layer,below=\step of l11.south,anchor=north] (add) {Add};
				\node[layer,below=\step of add.south,anchor=north]  (largeout) {Softmax};

				\draw[latex'-] (sh1.north) to ++(0,\step) node [above,inner sep=2pt,font=\footnotesize\strut] {Input: $32\times32\times3$};
				\draw[-latex'] (sh1)  to (sh2);
				\draw[-latex'] (sh2)  to (sh3);
				\draw[-latex'] (sh3)  to (sh4);
				\draw[-latex'] (sh4)  to (sh5);
				\draw[-latex'] (sh5)  to (sh6);
				\draw[-latex'] (sh6.south) -- ++(0,-0.3*\step) -| (l1.north);
				\draw[-latex'] (l1)   to (l2);
				\draw[-latex'] (l2)   to (l3);
				\draw[-latex'] (l3)   to (l4);
				\draw[-latex'] (l4)   to (l5);
				\draw[-latex'] (l5)   to (l6);
				\draw[-latex'] (l6)   to (l7);
				\draw[-latex'] (l7)   to (l8);
				\draw[-latex'] (l8)   to (l9);
				\draw[-latex'] (l9)   to (l10);
				\draw[-latex'] (l10)  to (l11);
				\draw[-latex'] (l11)  to (add);
				\draw[-latex'] (add)  to (largeout);
				\draw[-latex'] (largeout.south) to ++(0,-\step) node [right,anchor=north,font=\footnotesize\strut,inner sep=2pt,align=center] {Complex NN output\\[2pt]Parameters: 401,334\\Operations: 16,311,630};

				\node[layer,below right=\step and 1.7cm of sh6.south,anchor=north] (t1) {MaxPool $2{\times}2$};
				\node[layer,below=\step of t1.south,anchor=north]  (t2) {Dropout $0.5$};
				\node[layer,below=\step of t2.south,anchor=north]   (t3) {Dense $10$};
				\node[layer,below=\step of t3.south,anchor=north]   (t4) {Softmax};

				\draw[-latex'] (sh6.south) -- ++(0,-0.3*\step) -| (t1.north);
				\draw[-latex'] (t1)   to (t2);
				\draw[-latex'] (t2)   to (t3);
				\draw[-latex'] (t3)   to (t4);
				\draw[-latex'] (t3.south) |- ++(-1.7cm,-0.3*\step) |- ([yshift=0.7*\step]add.north) to (add);
				\draw[-latex'] (t4.south) to (t4.south |- largeout.south) to ++(0,-\step) node [right,anchor=north,font=\footnotesize\strut,inner sep=2pt,align=center] {Lightweight NN output\\[2pt]Parameters: 25,834\\Operations: 9,317,972};
			\end{tikzpicture}
		};
		\node[anchor=north,inner sep=1pt] at (sharefirst.south |- cascading.south) {a) Share first layers};
		\node[anchor=north,inner sep=1pt] at (cascading.south) {b) Cascading};
	\end{tikzpicture}
	\caption{
		Topologies of the neural networks for CIFAR-10.
		Cascading requires employing an additional branch dropout layer to maintain a high accuracy on the lightweight output.}
	\label{fig:cifar10_topo}
\end{figure}

\cref{fig:cifar10_topo}a and \cref{fig:cifar10_topo}b summarize the topologies for the two studied methods of parameter sharing, i.e., sharing only the first few layers, and a cascaded model which allows sharing of all parameters of the lightweight model.
We base the complex \ac{NN} on \cite{cifar10_large}.
In both cases, the lightweight \ac{NN} is identical.
In the case of sharing only the first few layers (\cref{fig:cifar10_topo}a), the two branches are independent except for the first six layers.
The lightweight \ac{NN} has fewer layers compared to the complex \ac{NN}, which reduces the number of parameters and computational operations to 6.7\,\% and 57.2\,\%, respectively.
The number of parameters and computational operations increases slightly in the case of cascading (\cref{fig:cifar10_topo}a), as the complex \ac{NN} now contains the entire lightweight \ac{NN}.
Adding the outputs of the last dense layers (before the softmax activation function) essentially results in a large dense layer for the complex \ac{NN}, from which only parts are also used by the lightweight \ac{NN}.
Powerful devices train both branches, whereas weak devices train only the lightweight branch.

When the powerful device trains both branches, it is possible that co-adaptations are created, in which each branch depends on the other branch to correct its mistakes.
The reason is that during training, the parameters of each branch are adjusted to reduce the loss given the output of the other branch.
Therefore, one branch may learn to correct the mistakes of the other branch~\cite{dropout}.
Generally, co-adaptations harm the accuracy of the \ac{NN} because they lead to overfitting~\cite{dropout}.
In our case, there is another issue with such co-adaptations: both branches are not always used together.
The lightweight branch must be able to work on its own when employed by weak devices.
Therefore, we add an additional dropout layer (branch dropout) to the end of the complex branch.
Whereas conventional dropout drops each neuron independently with a certain probability~\cite{dropout}, the branch dropout drops the whole branch with a certain probability (``all or nothing'').
The effect of this operation is that the lightweight branch cannot rely on the output of the complex branch and it, therefore, maintains a high accuracy if used in isolation.
We do not apply branch dropout to the lightweight branch because the complex branch is never used alone.
We purposefully do allow adaptations of the complex branch to the output of the weak branch to improve/correct its output.

We use \emph{RMSProp}~\cite{rmsprop} with learning rate $\eta=0.0001$ with decay of $10^{-6}$.
We perform training with mini-batches of size 32.
We base our implementation on the \emph{tensorflow.keras}~\cite{tensorflow} library.

\end{document}